\documentclass[conference]{IEEEtran}
\usepackage{cite}
\usepackage{amsmath,amssymb,amsfonts}
\usepackage{algorithmic}
\usepackage{graphicx}
\usepackage{textcomp}
\usepackage{xcolor}
\usepackage{flushend} %

\usepackage{bm}
\usepackage{bbm}

\usepackage{siunitx}

\definecolor{rose_color}{RGB}{219, 48, 122}
\definecolor{blue_color}{RGB}{72, 159, 240}
\definecolor{green_color}{RGB}{56, 232, 79}
\definecolor{grey_color}{RGB}{50, 76, 80}  %
\colorlet{orange_color}{orange}

\colorlet{los_color}{rose_color}
\colorlet{rup_color}{blue_color}
\colorlet{rdw_color}{green_color}
\colorlet{tx_color}{grey_color}
\colorlet{concat_color}{grey_color!20}
\colorlet{e3_color}{green_color}
\colorlet{op_color}{grey_color!20}
\colorlet{flow_color}{orange}
\colorlet{orange_color}{orange}

\colorlet{tx_color}{white}
\colorlet{rx_color}{white}

\usepackage{tikz}
\usetikzlibrary{shapes.geometric}
\usepackage{pgf}
\usepackage{pgfplots}
\pgfplotsset{compat=1.18}

\usepackage{import}  %

\tikzstyle{block}=[draw=black, thick, align=center, rounded corners, minimum height=0.5cm, fill=blue_color, text=black]
\tikzstyle{MLP}=[trapezium, trapezium angle=80, draw=black, thick, align=center, fill=blue!70, text=black, shape border rotate=-180, inner sep=3pt, outer sep=0pt]
\tikzstyle{state}=[block, fill=orange]
\tikzstyle{flow}=[font=\footnotesize, sloped, fill=none, inner sep=0pt, draw=none, minimum size=4mm]
\tikzstyle{invalid}=[opacity=0.3]
\tikzstyle{input}=[inner sep=0pt, align=center]
\tikzstyle{output}=[input]
\tikzstyle{concat}=[block, fill=concat_color]
\tikzstyle{op}=[circle, draw=black, thick, inner sep=1pt, fill=op_color, text=black, font=\bfseries]

\pgfdeclarelayer{bg}    %
\pgfsetlayers{bg,main}  %

\usepackage{url}
\usepackage{hyperref}
\usepackage{fontawesome5}

\definecolor{orcid_color}{HTML}{A6CE39}
\newcommand{\orcid}[1]{\href{https://orcid.org/#1}{{\color{orcid_color}\faOrcid{}} #1}}

\usepackage{glossaries-extra}

\setabbreviationstyle[acronym]{long-short}

\glssetcategoryattribute{acronym}{nohyperfirst}{true}

\newacronym{p2p}{P2P}{Point-to-Point}
\newacronym{rt}{RT}{Ray Tracing}
\newacronym{rl}{RL}{Ray Launching}
\newacronym{dynrt}{DynRT}{Dynamic \gls{rt}}
\newacronym{diffrt}{DiffRT}{Differentiable \gls{rt}}
\newacronym{go}{GO}{Geometrical Optics}
\newacronym{utd}{UTD}{Uniform Theory of Diffraction}
\newacronym{los}{LOS}{line of sight}
\newacronym{em}{EM}{electromagnetic}
\newacronym{mlp}{MLP}{Multi Layer Perceptron}
\newacronym{tx}{TX}{transmitter}
\newacronym{rx}{RX}{receiver}
\newacronym{ad}{AD}{automatic differentiation}
\newacronym{ml}{ML}{Machine Learning}
\newacronym{dag}{DAG}{Directed Acyclic Graph}
\newacronym{gflownets}{GFlowNets}{Generative Flow Networks}
\newacronym{2d}{2D}{two-dimensional}
\newacronym{3d}{3D}{three dimensions}

\def\BibTeX{{\rm B\kern-.05em{\sc i\kern-.025em b}\kern-.08em
T\kern-.1667em\lower.7ex\hbox{E}\kern-.125emX}}

\usepackage{fancyhdr}
\begin{document}

\title{Towards Generative Ray Path Sampling\\ for Faster Point-to-Point Ray Tracing}

\author{\IEEEauthorblockN{Jérome Eertmans}
  \IEEEauthorblockA{\textit{ICTEAM} \\
    \textit{Université catholique de Louvain} \\
    Louvain-la-Neuve, Belgium \\
  \orcid{0000-0002-5579-5360}}
  \and
  \IEEEauthorblockN{Nicola Di Cicco}
  \IEEEauthorblockA{\textit{DEIB} \\
    \textit{Politecnico di Milano} \\
    Milano, Italy \\
  \orcid{0000-0002-1524-8178}}
  \and
  \IEEEauthorblockN{Claude Oestges}
  \IEEEauthorblockA{\textit{ICTEAM} \\
    \textit{Université catholique de Louvain} \\
    Louvain-la-Neuve, Belgium \\
  \orcid{0000-0002-0902-4565}}
  \and
  \IEEEauthorblockN{Laurent Jacques}
  \IEEEauthorblockA{\textit{ICTEAM} \\
    \textit{Université catholique de Louvain} \\
    Louvain-la-Neuve, Belgium \\
  \orcid{0000-0002-6261-0328}}
  \and
  \IEEEauthorblockN{Enrico Maria Vitucci}
  \IEEEauthorblockA{\textit{DEI} \\
    \textit{University of Bologna} \\
    Bologna, Italy \\
  \orcid{0000-0003-4582-0953}}
  \and
  \IEEEauthorblockN{Vittorio Degli-Esposti}
  \IEEEauthorblockA{\textit{DEI} \\
    \textit{University of Bologna} \\
    Bologna, Italy \\
  \orcid{0000-0001-6589-8243}}
}

\maketitle

\thispagestyle{fancy}

\begin{abstract}
  Radio propagation modeling is essential in telecommunication research, as radio channels result from complex interactions with environmental objects. Recently, Machine Learning has been attracting attention as a potential alternative to computationally demanding tools, like Ray Tracing, which can model these interactions in detail. However, existing Machine Learning approaches often attempt to learn directly specific channel characteristics, such as the coverage map, making them highly specific to the frequency and material properties and unable to fully capture the underlying propagation mechanisms. Hence, Ray Tracing, particularly the Point-to-Point variant, remains popular to accurately identify all possible paths between transmitter and receiver nodes. Still, path identification is computationally intensive because the number of paths to be tested grows exponentially while only a small fraction is valid. In this paper, we propose a Machine Learning-aided Ray Tracing approach to efficiently sample potential ray paths, significantly reducing the computational load while maintaining high accuracy. Our model dynamically learns to prioritize potentially valid paths among all possible paths and scales linearly with scene complexity. Unlike recent alternatives, our approach is invariant with translation, scaling, or rotation of the geometry, and avoids dependency on specific environment characteristics.
\end{abstract}


\section{Introduction}
Over the past decade, \gls{rt} has emerged as a leading tool in the field of radio propagation modeling, offering a favorable balance between accuracy, ease of implementation, and computational efficiency \cite{imagemethod,visibilitytree}.

In \gls{p2p} \gls{rt}, e.g., Image-based \gls{rt}, the computational complexity is most impacted by the size of the scene or the number of interactions to be taken into account. Indeed, the number of path candidates, i.e., the \emph{potential} ray paths to be considered, grows more than linearly with respect to both parameters.

In recent years, numerous papers have investigated the potential applications of \gls{ml} to the modeling of radio propagation. Motivated by the need for reducing the computation time, \gls{ml} has been primarily utilized for two purposes: (1) the estimation of radio materials from measurements data or RT simulations, or scene reconstruction to later simulate radio propagation\cite{scenereconstruct,radioenvlearning}; and (2) the learning of wave propagation, based on measurements or simulations, in order to create a less expensive model that can make new predictions when the environment varies \cite{mlpropmodel,radiomapconvgan,digitaltwin,convdeep,wirelesspred,uav,losprob,winert}. For the second purpose, which is closely related to our application goal, most models try to learn either a specific, usually fixed-size, input scene, or the significantly complex relationship between some input geometries and the \gls{em} field. Since \gls{em} fields usually vary with the frequency, the radio materials, and numerous other parameters, these \gls{ml} models are constrained to operate within a specific frequency range or on a particular scene. Consequently, they are unable to generalize to other scenarios. Moreover, those models are usually trained to predict a specific channel characteristic, like the coverage map, and they cannot be reused to estimate other parameters. We hypothesize that \gls{ml} models are not optimal for the complete replacement of channel modeling tools such as \gls{rt}. This is particularly true given that the calculation of \gls{em} fields is usually straightforward once the ray paths are known. Instead, we believe that \gls{ml} is best suited for the learning of the most complex parts of the \gls{rt} pipeline, and should be used alongside already existing modeling tools, not as a complete replacement.

In this paper, we propose to address the issue of computational complexity by using an \gls{ml} model that learns how to identify only the valid ray paths that connect two nodes in a scene, typically a \gls{tx} and a \gls{rx}, and that have a given number of interactions with the environment. A ray path is considered to be valid if (1) it fulfills Fermat's principle and (2) it is not obstructed by any other object in the scene.

We summarize our major contributions as follows.
\begin{enumerate}
  \item We propose a \textbf{methodology for sampling path candidates based on generative \gls{ml}}, the first in the field to the best of our knowledge;
  \item We develop a \textbf{radio frequency- and material-independent} model that can be trained using reinforcement learning, thus avoiding the need to generate a---potentially computationally expensive---ground truth dataset;
  \item We implement an \gls{ml} model that is invariant under the special Euclidean group $\mathrm{SE}(3)$ in \gls{3d}, composed of the rotation ($\mathrm{SO}(3)$) and the translation groups, but also to arbitrary scaling, and that can support input scenes of \textbf{any size},
    which enables generalization to novel scenes unseen at training time.
\end{enumerate}

The remainder of this paper is organized as follows. \autoref{sec:motivations} motivates the use of a sampling-based approach to reduce the overall simulation time, and introduces the necessary notations. Then, \autoref{sec:related} discusses the recent applications of \gls{ml} to radio propagation. Subsequently, \autoref{sec:model} presents our model in a general framework, elucidating its operating principle, and \autoref{sec:application} demonstrates its efficiency through a \gls{3d} \gls{rt} example and preliminary results. Finally, \autoref{sec:conclusion} reiterates the contributions and preliminary outcomes of the model, while also outlining future avenues for research. The source code\footnote{Repository: \url{https://github.com/jeertmans/DiffeRT}.} and a comprehensive tutorial\footnote{Tutorial: \url{https://differt.rtfd.io/icmlcn2025/notebooks/sampling_paths.html}.} are also made available to guide readers through the simulation procedure.

\section{Motivations and Notations}\label{sec:motivations}

\begin{figure}
  \centering
  \begin{tikzpicture}  %
    \node[block, label=above:{\hypertarget{step:1}{(1)}}] (gp) at (0,0) {Geometry\\pre-processing};
    \node[block, text=black, fill=orange, rounded corners, label=above:{\hypertarget{step:2}{(2)}}] (pc) at (2.5,0) {Paths\\candidates};
    \node[block, label=left:{\hypertarget{step:3}{(3)}}] (pt) at (2.5,-1.75) {Paths\\tracing};
    \node[block, label=below:{\hypertarget{step:4}{(4)}}] (pp) at (2.5,-3.5) {Paths\\post-processing};
    \node[block, label=above:{\hypertarget{step:5}{(5)}}] (em) at (0,-3.5) {\gls{em} fields};
    \draw[<-,thick, rounded corners] (gp.west) -| ++(-.5,-1) node[below, align=center] {Input scene\\(\gls{tx}, \gls{rx}, objects, ...)};
    \draw[->, thick] (gp.east) -- (pc.west);
    \draw[->, thick] (pc.south) -- (pt.north);
    \draw[->, thick] (pt.south) -- (pp.north);
    \draw[->, thick] (pp.west) -- (em.east) node[midway] (midway) {};
    \draw[->, thick, rounded corners] (midway.center) |- ([yshift=-.5cm]em.south) node[left, align=center] {All rays from\\\gls{tx} to \gls{rx}};
    \draw[->, thick] (em.west) -- ++(-.5, 0) node[left, align=center] {\gls{em} field at\\\gls{rx}};
\end{tikzpicture}
  \caption{The pipeline of a typical \gls{p2p} \gls{rt} software where relevant steps are numbered from \protect\hyperlink{step:1}{1} to \protect\hyperlink{step:5}{5}. We propose to replace step \protect\hyperlink{step:2}{2} with an \gls{ml} model.}
  \label{fig:pipeline}
\end{figure}

\begin{figure}
  \centering
  \import{pgf}{many_path_candidates.pgf}
  \caption{Illustration of the many path candidates vs. few valid paths issue in a 2D scene with blue obstacles. Here, only \SI{10}{\%} of all path candidates form valid ray paths (orange solid lines) between \gls{tx} and \gls{rx}.}
  \label{fig:many_paths}
\end{figure}

In the majority of \gls{p2p} \gls{rt} frameworks, the process of estimating the \gls{em} fields in a given scene can be summarized as outlined in \autoref{fig:pipeline}. It should be noted that the computational complexity of individual steps varies, with paths computation (steps \hyperlink{step:2}{2} to \hyperlink{step:4}{4} in \autoref{fig:pipeline}) being particularly demanding in the context of large scenes or higher order paths.

Specifically, let the size of the scene, i.e., the number of objects, be $N$, and the number of interactions be the variable $K$, then the number of possible path candidates (step \hyperlink{step:2}{2}) is at most $N^K$. A path candidate is simply an ordered sequence, of length $K$, indicating the objects with which the ray path interacts. Then, for each of the path candidates, an actual geometric ray path is constructed (step \hyperlink{step:3}{3}), e.g., using the image method or other methods, such as the minimization-based ones \cite{imagemethod,fpt,mpt}. Constructing a ray path from a path candidate usually takes a time proportional to the number of interactions, i.e., $\mathcal{O}\left(K\right)$. After that, each ray path is checked (step \hyperlink{step:4}{4}) to see if it is an actual valid path or not. This check is usually bounded by $\mathcal{O}\left(N K\right)$ operations, assuming that checks are performed for each ray path segment. Finally, the number of ray paths used to compute \gls{em} fields (step \hyperlink{step:5}{5}) is usually orders of magnitude smaller than the number of possible path candidates (see \autoref{fig:many_paths}).

In city-scale \gls{3d} scenes, $N$ can easily exceed $10^6$ (e.g., the number of triangular facets), and $N^K$ thus becomes too large either to fit in memory or to be simulated in a reasonable time. There are three principal countermeasures to this fundamental problem: (i) resort to \gls{rl}, a method that launches rays with a given angular separation from a transmitter node, e.g., \gls{tx}, and collects all rays passing in the vicinity of a receiver node, e.g., \gls{rx}; (ii) use a heuristic approach to reduce the number of path candidates (e.g., the \textit{Fibonacci} method used in Sionna \cite{sionnart}); or (iii) compute some kind of visibility tree---or visibility matrix---to avoid generating path candidates that eventually lead to invalid ray paths\cite{visibilitytree,mpt}. Solution (iii) is potentially time-efficient; however, its implementation in practice is challenging, since computing the visibility in a \gls{3d} scene is itself computationally expensive. Solution (ii) is a hybrid of \gls{rl} and \gls{p2p} \gls{rt}, and may depend on how good the heuristic is.

\Gls{ml} models have already shown very impressive results in generating discrete objects such as graphs \cite{zhu2022a} and protein structures \cite{jumper2021highly}. Following this line of investigation, instead of relying on a heuristic such as solution (ii), our motivation is to construct an \gls{ml} model that will learn how to sample valid path candidates in the set of all possible path candidates (step \hyperlink{step:2}{2}).
In essence, our objective is to learn an \gls{ml} model that constructs a probability-based visibility tree, similarly to solution (iii).

\section{Related Work}\label{sec:related}

\Gls{ml} methods have already been applied to a variety of radio propagation applications, frequently in combination with \gls{rt}, most of the time to train the model by generating data. For example, \gls{ml} has been used to predict the propagation of electromagnetic waves in wireless networks \cite{mlpropmodel,radiomapconvgan,digitaltwin,convdeep,wirelesspred,uav,losprob,winert,winertextended}. In many of these cases, the \gls{ml} model will eventually replace the entire channel modeling tool, thereby eliminating the ability to obtain any intermediate, interpretable outputs, such as those provided by \gls{rt} in the form of geometrical ray paths. Furthermore, we have identified the following limitations in the existing models: (i) they often necessitate the acquisition of measurements or many simulations for training, (ii) they depend on a specific frequency and the radio materials from the scene, and, most importantly, (iii) they are typically trained on a fixed scene and cannot generalize to arbitrarily-sized inputs.

As previously stated, we posit that \gls{ml} is ill-suited to predict the final outputs of \gls{rt} tools, such as coverage maps, as they are often chaotic and depend on an excessive number of parameters. Instead, we propose the creation of an \gls{ml}-assisted \gls{rt} pipeline that employs the \gls{ml} model solely to reduce the overall computational complexity, rather than the deterministic and computationally simple procedures, such as \gls{em} coefficients, which can be calculated with certainty once a ray path is known. To the best of our knowledge, the WiNeRT model \cite{winert} is the only attempt to address the challenge of path tracing by learning the relationship between incident and outgoing rays for each interaction, e.g., reflections, as well as the \gls{em} propagation.

In comparison to existing literature, our model offers the following main novelties:
\begin{enumerate}
  \item To the best of our knowledge, our model is the first to use \gls{ml} to sample path candidates;
  \item Because our model only samples path candidates, it is therefore radio-material- and frequency-independent;
  \item It is capable of accommodating input scenes of any size due to the manner in which the input scene is transformed, as detailed in the subsequent section.
  \item Furthermore, the model does not require any a priori ground truth training set and can be fully trained using any ray tracer, while completely ignoring the \gls{em} fields.
\end{enumerate}

The subsequent section will present the \gls{ml} model architecture that incorporates all the aforementioned properties.

\section{Methodology}\label{sec:model}

\begin{figure*}
  \centering
  \begin{tikzpicture}
  [level distance=18mm,
   every node/.style={state},
   edge from parent/.style={draw, rounded corners, thick},
   level 1/.style={sibling distance=60mm},
   level 2/.style={sibling distance=20mm}]
  \node {??}
     child {node {A?}
       child {node[invalid] {AA}
            edge from parent[dashed] node[flow, above] {$F(\text{A?},\text{AA})$}
       }
       child {node {AB}
            edge from parent node[flow, fill=white, rotate=90] {$F(\text{A?},\text{AB})$}
       }
       child {node {AC}
            edge from parent node[flow, above] {$F(\text{A?},\text{AC})$}
       }
        edge from parent node[flow, above] {$F(\text{??},\text{A?})$}
     }
     child {node {B?}
       child {node {BA}
            edge from parent node[flow, above] {$F(\text{B?},\text{BA})$}
       }
       child {node[invalid] {BB}
            edge from parent[dashed] node[flow, fill=white, rotate=90] {$F(\text{B?},\text{BB})$}
       }
       child {node {BC}
            edge from parent node[flow, above] {$F(\text{B?},\text{BC})$}
       }
        edge from parent node[flow, fill=white, rotate=90] {$F(\text{??},\text{B?})$}
     }
     child {node {C?}
       child {node {CA}
            edge from parent node[flow, above] {$F(\text{C?},\text{CA})$}
       }
       child {node {CB}
            edge from parent node[flow, fill=white, rotate=90] {$F(\text{C?},\text{CB})$}
       }
       child {node[invalid] {CC}
            edge from parent[dashed] node[flow, above] {$F(\text{C?},\text{CC})$}
       }
        edge from parent node[flow, above] {$F(\text{??},\text{C?})$}
     };
\end{tikzpicture}
  \caption{Flowchart of all possible states for $K=2$ path candidates in a scene with $N=3$ objects: A, B, and C. Each state $s$ can flow to three possible child states $s'$, to select interacting objects one at a time. As interacting with the same object twice in a row is \textbf{physically unsound}, this state is marked as \textbf{unreachable} (faded nodes and dotted lines). To account for that in the model, the flow is stopped, i.e., set to zero, to prevent reaching those states.}
  \label{fig:path_candidates}
\end{figure*}

As previously stated, the objective of our model is to serve as a surrogate for the conventional path candidates generation process (step \hyperlink{step:2}{2} in \autoref{fig:pipeline}). The objective is thus to learn a generative random function $f_w$, depending on TX, RX and the list of objects OBJECTS and parametrized by a set of learnable weights $w$, and such that each call to $f_w$ provides, with good probability, a valid path candidate. Moreover, the learning of $w$ is expected to maximize the probability  
\begin{equation}\label{eq:accuracy}
    \mathbb P\big[f_w(\text{TX}, \text{RX}, \text{OBJECTS}) = \text{VALID PATH}\big]. 
\end{equation}

To generate all paths candidates, we construct a \emph{search tree} (see \autoref{fig:path_candidates}), in that it is strictly related to the concept of visibility tree in \gls{rt} \cite{visibilitytree}. For a given number of interactions $K$, we can construct each possible path candidate by descending from the root node, also referred to as the initial state, and take one of the possible branches. Each branch corresponds to an object in the scene, identified here with a unique letter, and a path candidate is constructed by identifying a list of objects to visit. As a ray path cannot usually interact with the same object twice in a row, those branches should be \textbf{unreachable} (dotted lines). The question mark ``?'' is used as an indicator of a path candidate being incomplete.

\begin{figure*}
  \centering
  \begin{tikzpicture}[label distance=0mm]  %
    \node[input, anchor=east, label=above:{\tiny $3$}] (tx) at (0, +.7) {\Glsxtrshort{tx}};
    \node[input, anchor=east, label=above:{\tiny $3$}] (rx) at (0, 0) {\Glsxtrshort{rx}};
    \node[input, anchor=east, label=above:{\tiny $N \times V \times 3$}] (objects) at (0, -.7) {OBJECTS};

    \path (rx) -- ++(1, 0) node[concat, minimum width=1.8cm, rotate=90] (concatxyz) {Concat};
    \draw[->, thick] (tx) -- (concatxyz.north |- tx);
    \draw[->, thick] (rx) -- (concatxyz.north |- rx);
    \draw[->, thick] (objects) -- (concatxyz.north |- objects);

    \path (concatxyz) -- ++(1, 0) node[input, label=above:{\tiny $(2 + NV)\times 3$}] (xyz) {XYZ};
    \draw[->, thick] (concatxyz) -- (xyz);

    \path (xyz) -- ++(0.7, -1) node[block, fill=concat_color] (mean) {Mean};

    \path (xyz) -- ++(1.4, 0) node[op] (translate) {$\boldsymbol{-}$};
    \draw[->, thick, rounded corners] ([yshift=-1mm]xyz.south) |- (mean)
    (mean.east) -| (translate);
    \draw[->, thick] (xyz) -- (translate);

    \path (translate) -- ++(.5, 0) node[op, anchor=west] (scale) {$\boldsymbol{\div}$};
    \path (xyz) -- (scale) node[midway, block, fill=concat_color, shift={(0,1)}] (std) {Std. dev.};
    \draw[->, thick, rounded corners] ([yshift=4mm]xyz.north) |- (std) -| (scale);
    \draw[->, thick] (translate) -- (scale);

    \path (scale.east) -- ++(.4, 0) node[concat, anchor=north, minimum width=1.5cm, rotate=90] (select) {Select};
    \draw[->, thick] (scale) -- (select) node[midway] (midselect) {};

    \path (select) -- ++(.5, 0) node[input, anchor=west, label=above:{\tiny $N \times V \times 3$}] (feat) {OBJECTS};
    \draw[->, thick] (select) -- (feat);

    \path (feat) -- ++(.7, 1) node[MLP, fill=e3_color] (mlp_scene) {MLP};
    \path (mlp_scene.east) -- ++(.3, 0) node[input, anchor=west, label=above:{\tiny $d$}] (scene_feat) {SCENE};
    \draw[->, thick, rounded corners] (midselect.center) |- (mlp_scene) -- (scene_feat);

    \path (feat) -- ++(.7, -1) node[MLP, fill=e3_color] (mlp_obj) {MLP};
    \path (mlp_obj.east) -- ++(.3, 0) node[input, anchor=west, label=above:{\tiny $N \times d$}] (obj_feat) {OBJECTS};
    \draw[->, thick, rounded corners] ([yshift=-1mm]feat.south) |- (mlp_obj) -- (obj_feat);

    \path (feat.east) -- ++(3.4, 0) node[concat, minimum width=1.8cm, rotate=90] (concatfeat) {Concat};
    \path (concatfeat) -- ++(.5, 0) node[MLP, anchor=west, fill=flow_color] (mlp_flow) {MLP};
    \draw[->, thick, dashed] (concatfeat) -- (mlp_flow);

    \path (mlp_flow) -- ++(.7, 0) node[input, anchor=west, label=above:{\tiny $N$}] (flows) {FLOWS};
    \draw[->, thick, dashed] (mlp_flow) -- (flows);
    \draw[->, thick, rounded corners] (obj_feat.east) -- ++(0.2, 0) coordinate (tmp) |- (concatfeat.north |- objects);
    \draw[->, thick, rounded corners] (scene_feat.east) -- (tmp |- scene_feat) |- (concatfeat.north |- tx);

    \path (obj_feat.north |- feat) -- ++(0.3, 0) node[MLP, anchor=west, fill=flow_color] (cell_flow) {CELL};
    \draw[->, thick, rounded corners, dashed] ([yshift=3mm]obj_feat.north) |- (cell_flow) -- (concatfeat);

    \path (flows) -- ++(.8, 0) node[concat, anchor=north, minimum width=1.5cm, rotate=90] (sample) {Sample};
    \draw[->, thick, dashed] (flows) -- (sample);

    \path (sample.south) -- ++(0.2, 0) node[input, anchor=west, label=above:{\tiny scalar}] (obj_index) {OBJECT\\INDEX};
    \draw[->, thick, dashed] (sample) -- (obj_index);

    \path (obj_index.east) -- ++(0.2, 0) node[op, anchor=west] (append) {[]};
    \draw[->, thick, dashed] (obj_index) -- (append);

    \path (append.north) -- ++(0, 0.3) node[output, anchor=south, label=above:{\tiny $0\text{~to~}K$}] (pc) {PATH\\CAND.};
    \draw[->, thick, dashed] (append) -- (pc);
    \draw[->, thick, rounded corners, dashed] (append.east) -| ++(.3, -.5) -| (append.south) node[below, align=center, font=\tiny, label={[align=center, font=\tiny]below:{Repeat\\$K$\\times}}] {};

    \path (concatfeat) -- (sample) node[midway] (mid) {};
    \draw[<-, thick, rounded corners] (sample.west) |- ++(-.3, -.3) node[left] {random key};

    \node[input, label=above:{\tiny $K \times N$}] (state) at (mid |- scene_feat) {STATE};
    \draw[->, thick, rounded corners, dashed] (state.west) -| (cell_flow);

    \path (pc.west) -- (obj_index |- state) node[op, anchor=east] (onehot) {$\mathbbm{1}$};

    \draw[->, thick, rounded corners, dashed] (pc.west) to[out=180, in=0] (onehot.east);
    \draw[->, thick, dashed] (onehot.west) -- (state.east);

\end{tikzpicture}
  \caption{Simplified representation of the proposed \gls{ml} model replacing the second step in \autoref{fig:pipeline}, with trainable weights as colored trapezoids. Each state, one-hot-encoded (\(\mathbbm{1}\)) from the current path candidate (size \(0\) to \(K\)), records visited objects, starting from a zero matrix. A recurrent cell maps it to the feature space for positional encoding. Dashed lines indicate state updates, and randomness comes from using a different key per path candidate.}
  \label{fig:model}
\end{figure*}

Our approach leverages the principles of \gls{gflownets} \cite{gflownet} \gls{ml} for learning to efficiently traverse the search tree. Instead of relying on exhaustive search, we model the process of generating a path candidate, $\mathcal{P}$, as \emph{flowing} in a \gls{dag}, starting from an initial state, and ending at a terminal state, the latter representing a complete path candidate. Each child state $s'$ has one unique parent state $s$, e.g., state ``AB'' is a child state of ``A?''.

Each terminal state, i.e., a path candidate, is associated with a scalar reward $R$. Intuitively, greater reward values represent ``better'' paths. An example reward function could return 1 if the path candidate ends up generating a valid ray path, and 0 otherwise. We will discuss later how we can define a better reward function. The central property of \gls{gflownets} is that, after successful training, the model is expected to sample terminal states, i.e., complete path candidates, with a probability $p$ that is proportional to their corresponding reward, $R$. That is,
\begin{equation}
  p(\mathcal{P}) \propto R(\mathcal{P}).
\end{equation}

Given the example reward function, \gls{gflownets} will learn to sample from the distribution of valid paths only.

To achieve this, the model must be trained to respect the following fundamental properties:
\begin{enumerate}
  \item Each edge in the search graph must be assigned a positive \emph{flow}, $F(s,s') > 0$, where $s$ is the parent state and $s'$ is the child state;
  \item Flow conservation between ingoing and outgoing edges must be ensured:
    \begin{equation}\label{eq:flow_consistency}
      \forall s', F(s,s') = R(s') + \sum_{s''}F(s',s''),
    \end{equation}
    that is, the sum of output flows, $F(s',s'')$, must be equal to the input flow, $F(s,s')$, minus the reward;
  \item The probability of choosing state $s'$ given state $s$ must be defined as
    \begin{equation}\label{eq:proba}
      p(s'|s) = \frac{F(s,s')}{\sum_{s''}F(s,s'')},
    \end{equation}
    that is, the probability of traversing an edge in the search graph is equal to its flow value normalized over all outgoing edges.
\end{enumerate}

It should be noted that all non-terminal states are assigned a zero reward value. However, this is not a limitation of the model itself, and the potential benefits of integrating partial rewards in the learning process will be discussed in greater details later on.

As illustrated in \autoref{fig:model}, the proposed model is divided into two parts: the left part, depicted in green, is responsible for transforming the input scene into feature vectors, and the second part, represented in orange, generates the flow from a specified state and scene to all potential objects. The subsequent section will provide a more detailed analysis of the former part.

\subsection{Extracting Feature Vectors From a Scene}

The initial stage of the model is designed to extract feature vectors from the scene, including the positions of the \gls{tx} and \gls{rx} elements, as well as an arbitrary list of object coordinates. In \autoref{fig:model}, we assume \gls{3d} coordinates; however, our model is also applicable to any other dimension, such as 2D coordinates. The sole restriction on the input scene is that each object must be represented by a fixed number of vertices, denoted by $V$. To accommodate scenes of arbitrary size and order, our model is based on the Deep Sets architecture\cite{deepsets}. To ensure invariance with respect to both translation and scaling, a standard normalization of the input scene coordinates is performed. Finally, the model itself is built on the \texttt{E3x} \gls{ml} library \cite{e3x}, which in turn utilizes two \glspl{mlp} to convert XYZ coordinates into vectors of $d$ features, thereby guaranteeing invariance with respect to rotation.

\subsection{Flow and Loss Function}

The second part of our model generates, for a given state $s$, an array of flows to each object. Then, the next state $s'$ is chosen based on  those flows. The diversity in the generated path candidates is provided by the random sampling performed between each state, where each possible output state $s'$ is weighted according to its flow $F(s,s')$, see \eqref{eq:proba}.

In practice, the flow function as detailed previously also depends on two other parameters, the scene feature vector and the object features, but were omitted for readability.

For training our model, we minimize the \gls{gflownets} loss function, which rewrites \eqref{eq:flow_consistency} as a mean squared error:
\begin{equation}\label{eq:loss}
  L(s') = \left(F(s,s') - R(s') - \sum_{s''}F(s',s'') \right)^2.
\end{equation}

We summarize the training procedure as follows:
\begin{enumerate}
  \item Generate, for a given random scene, a batch of path candidates;
  \item Evaluate the \gls{gflownets} loss \eqref{eq:loss} for a batch of path candidates, and accumulate the results;
  \item Perform a gradient-based update of the \gls{ml} model, minimizing the \gls{gflownets} loss;
\end{enumerate}
and repeat those steps for a large number of scenes.

Upon successful training, as a consequence of the reward function, the model is expected to sample all valid path candidates with the same probability, with a computational cost scaling linearly with the number of objects, reflections, and model size: \(\mathcal{O}(cPNK)\), where \(c\) is a constant for model calls and \(P\) the number of sampled path candidates. For large scenes, this is significantly cheaper than the exhaustive approach, having a cost of \(\mathcal{O}(N^K)\).

\subsection{Accuracy and Hit Rate}

As our loss function may not directly reflect the model performance, we introduce two metrics to better evaluate its effectiveness: accuracy and hit rate. The \textbf{accuracy} is defined as the ratio of valid paths to the number of path candidates generated, approaching the probability of sampling a valid path defined in \eqref{eq:accuracy}. An accuracy of \SI{100}{\percent} indicates that the model exclusively samples path candidates that ultimately result in valid ray paths. The \textbf{hit rate} is defined as the number of valid generated ray paths divided by the total number of valid ray paths that could be generated. A hit rate of \SI{100}{\percent} means that the model samples all potential solutions.

\section{Application to a 3D Scenario}\label{sec:application}

In this section, we present an application of our model to sampling path candidates in the context of \gls{3d} \gls{p2p} \gls{rt} in an urban street canyon, with first and second order specular-reflection paths only. The complete procedure is available in the aforementioned tutorial notebook.

\subsection{Training Set}

\begin{figure}
  \centering
  \includegraphics[width=.6\columnwidth]{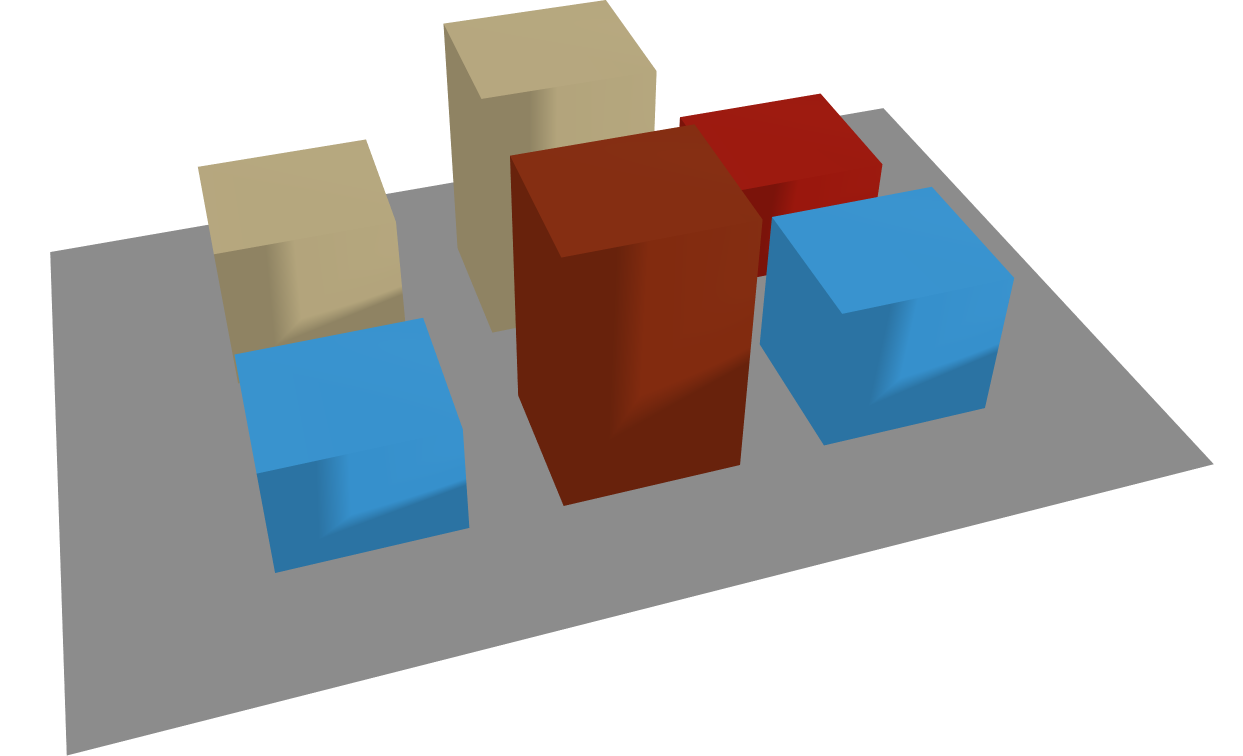}
  \caption{Street canyon scene from Sionna \cite{sionna}.}
  \label{fig:street_canyon}
\end{figure}

We generate our training set by randomly sampling modified versions of the ``\emph{simple street canyon}'' scene from Sionna \cite{sionnart} (see \autoref{fig:street_canyon}):
\begin{itemize}
  \item TX and RX are randomly positioned in the main street canyon;
  \item A random number of triangle facets is removed.
\end{itemize}
In each scene configuration, a number of path candidates are generated, and the corresponding ray paths are constructed. The validity of each generated path is determined using a standard ray-obstruction check, and a gradient-based update is performed to minimize the loss function \eqref{eq:loss}.

For validation, we generate 100 new scenes that we will use to measure the evolution of the accuracy and the hit rate of the model throughout the learning steps.

\subsection{Simulation Parameters}

The model was trained by performing \num{100e3} steps with the Adam optimizer \cite{adam} (learning rate of \num{3e-5}). For each step, the loss was accumulated by generating 100 path candidates. The number of features was set to $d=100$. The first two \glspl{mlp} have 3 linear layers and a hidden size matching the number of features. The last \gls{mlp} has 3 linear layers with a hidden size of 500. Finally, the model's accuracy and hit rate were evaluated by sampling 10 path candidates on each scene of the validation set.

\subsection{Results}

\DeclareSIUnit{\nothing}{\relax}

\begin{figure}
  \centering
  \begin{tikzpicture}[trim axis left, trim axis right]
    \pgfplotsset{
      width=0.9\columnwidth,
      height=3.95cm,
      scaled x ticks=false,
      scaled y ticks=false,
    }
    \begin{axis}[
        name=k1,
        axis y line*=left,
        xticklabels=none,
        legend pos=south east,
      ]
      \addlegendimage{empty legend}\addlegendentry{$K=1$}
      \addplot +[
        draw=rose_color,
        mark=none,
      ] table [x=steps, y expr=\thisrow{acc1} * 100] {data/training.txt} node[pos=0.51,yshift=-.20cm] (midacc1) {};
      \draw[<-, rose_color] (midacc1) -- ++(-45:.5cm) node[anchor=north west, font=\small, inner sep=0pt] {Accuracy};
    \end{axis}
    \begin{axis}[
        axis y line*=right,
        axis x line=none,
      ]
      \addplot +[
        draw=blue_color,
        mark=none,
        dashed,
      ] table [x=steps, y expr=\thisrow{hr1} * 100] {data/training.txt};
    \end{axis}
    \begin{axis}[
        name=k2,
        at={(k1.south)},
        yshift=-.2cm,
        anchor=north,
        xlabel = {Training steps},
        xlabel near ticks,
        xlabel style={inner sep=0pt},
        xtick = {0, 25e3, 50e3, 75e3, 100e3},
        xticklabels = {0, \SI{25}{\kilo\nothing}, \SI{50}{\kilo\nothing}, \SI{75}{\kilo\nothing}, \SI{100}{\kilo\nothing}},
        axis y line*=left,
        legend pos=south east,
      ]
      \addlegendimage{empty legend}\addlegendentry{$K=2$}
      \addplot +[
        draw=rose_color,
        mark=none,
      ] table [x=steps, y expr=\thisrow{acc2} * 100] {data/training.txt};
    \end{axis}
    \begin{axis}[
        at={(k2.north)},
        anchor=north,
        axis y line*=right,
        axis x line=none,
      ]
      \addplot +[
        draw=blue_color,
        mark=none,
        dashed,
      ] table [x=steps, y expr=\thisrow{hr2} * 100] {data/training.txt} node[pos=.21,yshift=+.08cm] (midhr2) {};
      \draw[<-, blue_color] (midhr2) -- ++(+135:.5cm) node[anchor=south east, font=\small, inner sep=1pt] {Hit rate};
    \end{axis}
    \path (k1.south west) -- ++(-1, 0) node[rotate=90] {Accuracy (\si{\percent})};
    \path (k1.south east) -- ++(+1, 0) node[rotate=-90] {Hit rate (\si{\percent})};
  \end{tikzpicture}
  \caption{Evolution of the accuracy and the hit rate throughout training steps.}
  \label{fig:training}
\end{figure}

\autoref{fig:training} compares the evolution of the accuracy and the hit rate throughout the training steps, for single-bounce reflection ($K=1$) and double-bounce reflection ($K=2$). First, we observe a short plateau in the learning curves for the first learning steps of the single-bounce model, where it starts with the same accuracy as a random sampler (\SI{3}{\percent}). The double-bounce model is initialized from the single-bounce model, after training, which explains a higher starting accuracy than a random sampler (\SI{0.03}{\percent}). Finally, a clear relationship can be observed between the hit rate and the accuracy, which shows that the model learns to generate different valid paths, with an increasing proportion of sampled paths being valid.

\subsection{Discussion}

The results are encouraging, particularly given the compact model size, when compared with random sampling performance. For $K=1$, both accuracy and hit rate steadily improve, indicating that the model effectively learns to prioritize valid paths. With an average hit rate of nearly \SI{80}{\percent}, the model identifies more than three-quarters of all possible valid paths by sampling only 10 candidates out of an average of 73. Although promising, this performance remains insufficient to fully replace exhaustive search, requiring further refinements for real-world applications.

For $K=2$, the model initially surpasses random sampling but shows only marginal improvements during training before stabilizing. Training from scratch without initializing from $K=1$ leads to collapse, with accuracy and hit rate dropping to zero, suggesting misalignment between the loss function and model performance. Higher $K$ values further exacerbate this issue due to sparse rewards from the low proportion of valid paths. Alternative reward functions, such as those inversely proportional to path length, have accelerated convergence but typically yield lower final performance. 

The current \gls{ml} model consists of three small \glspl{mlp}, enabling rapid sampling but potentially limiting scalability to complex scenes. Increasing layer depth could improve performance in larger environments, while expanding feature vector size has already shown benefits, albeit at the cost of slower training.

\section{Conclusion}\label{sec:conclusion}

This work presents a novel \gls{ml} model that learns how to sample paths. When properly trained, it could potentially remove the exponential complexity behind the usual \gls{p2p} \gls{rt} toolboxes. The model learns how to prioritize some branches in the complex search tree that represents the set of all possible path candidates. Our results demonstrate that, particularly in the single-bounce case ($K=1$), the model effectively learns to generate valid paths across diverse scenes. However, for more complex multi-bounce scenarios ($K \geq 2$), our preliminary results indicate that additional improvements are necessary.

Future work will explore several key areas to enhance the model's performance. First, addressing sparse reward functions, potentially through reward smoothing or continuous reward formulations \cite{fully-eucap2024,sparsegenerative,metagsparse}, may aid in learning across more complex scenes. Additionally, scaling the model architecture by increasing layer size and depth could improve handling of intricate environments. Investigating alternative training approaches, such as pre-training on simpler scenes before scaling up, may also prove beneficial. Ultimately, our goal is to deploy this model in larger, more complex \gls{3d} urban scenes to substantively reduce computational demands, particularly within \gls{rt} tools like Sionna or our own differentiable Ray Tracer, DiffeRT\footnote{Repository: \url{https://github.com/jeertmans/DiffeRT}.}, where complexity reduction could yield substantial computational savings.

\bibliographystyle{IEEEtran}
\bibliography{IEEEabrv,biblio}

\end{document}